\documentclass[conference]{IEEEtran}
\usepackage{amsmath}
\usepackage{amsfonts}
\usepackage{multirow}
\usepackage{graphicx}
\usepackage{footnote}
\usepackage{times}
\usepackage{textcomp}
\makesavenoteenv{tabular}
\makesavenoteenv{table}
\usepackage{algorithmicx}
\usepackage{algpseudocode}
\usepackage{algorithm}
\usepackage{cite}

\ifCLASSOPTIONcompsoc
  \usepackage[caption=false,font=normalsize,labelfont=sf,textfont=sf]{subfig}
\else
  \usepackage[caption=false,font=footnotesize]{subfig}
\fi

\ifCLASSINFOpdf
\else
\fi
\begin{document}
\title{Is Big Data Sufficient for a Reliable Detection of Non-Technical Losses?}

%\author{\IEEEauthorblockN{Patrick Glauner, Andre Boechat, Lautaro Dolberg, Radu State,}
%\IEEEauthorblockA{Interdisciplinary Centre for Security, Reliability and Trust, \\
%University of Luxembourg\\
%2721 Luxembourg, Luxembourg\\
%Email: \{first.last\}@uni.lu}
%\and
%\IEEEauthorblockN{Franck Bettinger, Yves Rangoni and Diogo Duarte}
%\IEEEauthorblockA{Choice Technologies Holding S\`arl\\
%2-4, rue Eug\`ene Ruppert \\
%2453 Luxembourg, Luxembourg \\
%Email: \{first.last\}@choiceholding.com}
%}


\author{\IEEEauthorblockN{Patrick Glauner\IEEEauthorrefmark{1},
Angelo Migliosi\IEEEauthorrefmark{1},
Jorge Augusto Meira\IEEEauthorrefmark{1},
Petko Valtchev\IEEEauthorrefmark{1}\IEEEauthorrefmark{2},
Radu State\IEEEauthorrefmark{1} and
Franck Bettinger\IEEEauthorrefmark{3}}
\IEEEauthorblockA{\IEEEauthorrefmark{1}Interdisciplinary Centre for Security, Reliability and Trust, University of Luxembourg\\
2721 Luxembourg, Luxembourg\\
Email: \{first.last, jorge.meira\}@uni.lu}
\IEEEauthorblockA{\IEEEauthorrefmark{2}Department of Computer Science, University of Quebec in Montreal, \\
H3C 3P8 Montreal, Canada\\
Email: valtchev.petko@uqam.ca}
\IEEEauthorblockA{\IEEEauthorrefmark{3}CHOICE Technologies Holding S\`arl\\
2453 Luxembourg, Luxembourg \\
Email: franck.bettinger@choiceholding.com}
}

\IEEEoverridecommandlockouts\IEEEpubid{\makebox[\columnwidth]{978-1-5090-4000-1/17/\$31.00~\copyright 2017 IEEE \hfill} \hspace{\columnsep}\makebox[\columnwidth]{ }}


\maketitle

\begin{abstract}
Non-technical losses (NTL) occur during the distribution of electricity in power grids and include, but are not limited to, electricity theft and faulty meters. In emerging countries, they may range up to 40\% of the total electricity distributed. In order to detect NTLs, machine learning methods are used that learn irregular consumption patterns from customer data and inspection results. The Big Data paradigm followed in modern machine learning reflects the desire of deriving better conclusions from simply analyzing more data, without the necessity of looking at theory and models. 
However, the sample of inspected customers may be biased, i.e. it does not represent the population of all customers.
As a consequence, machine learning models trained on these inspection results are biased as well and therefore lead to unreliable predictions of whether customers cause NTL or not. In machine learning, this issue is called covariate shift and has not been addressed in the literature on NTL detection yet.
In this work, we present a novel framework for quantifying and visualizing covariate shift. We apply it to a commercial data set from Brazil that consists of 3.6M customers and 820K inspection results. 
We show that some features have a stronger covariate shift than others, making predictions less reliable. In particular, previous inspections were focused on certain neighborhoods or customer classes and that they were not sufficiently spread among the population of customers.
This framework is about to be deployed in a commercial product for NTL detection.
%Furthermore, the observations made in this contribution will lay the foundations for future work on making NTL predictors more reliable.
\end{abstract}


% \fi
\IEEEpeerreviewmaketitle

\section{Introduction}
Losses in power grids can be grouped into technical losses and non-technical losses. Technical losses occur naturally, which are mainly caused by internal electrical resistance of infrastructure components. Non-technical losses (NTL) appear during power distribution and include, but are not limited to, the following causes \cite{chauhan2013non, glauner2016challenge}:
\begin{itemize}
\item Meter tampering in order to record lower consumptions
\item Bypassing meters by rigging lines from the power source
\item Arranged false meter readings by bribing meter readers
\item Faulty or broken meters
\item Technical and human errors in meter readings, data processing and billing
\end{itemize}

In practice, NTLs primarily consist of electricity theft and cause major problems to electricity providers, including financial losses and a decrease of stability and reliability. Furthermore, NTLs lead to an extra use of limited natural resources which in turn increases pollution. They can range up to 40\% of the total electricity distributed in countries such as Brazil, India, Malaysia or Lebanon \cite{depuru2013high, nagi2011improving}.

The predominant research direction reported in the recent literature is the use of machine learning/data mining methods, which learn anomalous behavior from customer data and known irregular behavior that was reported through inspection results.
However, carrying out inspections is expensive, as it requires physical presence of technicians. It is therefore important that the trained models make accurate predictions.

\begin{figure}[!t]
\centering
\includegraphics[width=0.37\textwidth]{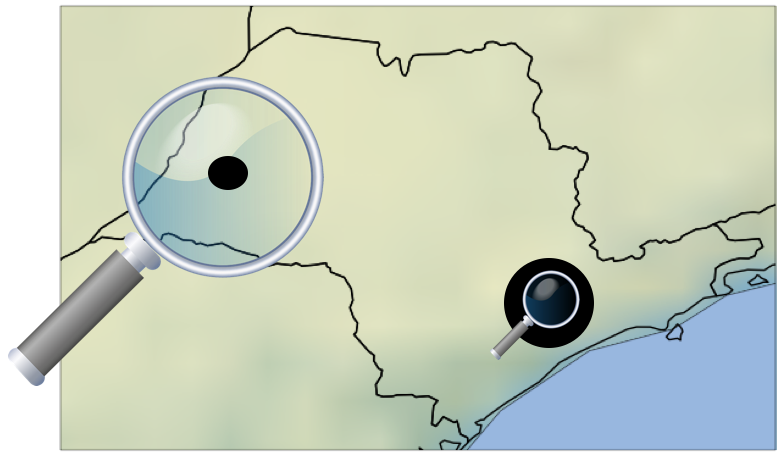}
\caption[XXX]{Example of spatial bias: The large city is close to the sea, whereas the small city is located in the interior of the country. The weather in the small city undergoes stronger changes during the year. The subsequent change of electricity consumption during the year triggers many inspections. As a consequence, most inspections are carried out in the small city. Therefore, the sample of customers inspected does not represent the overall population of customers.}
\label{fig:example}
\end{figure}

For about the last fifteen years, the Big Data paradigm followed in machine learning has been to gather more data rather than improving models. Hence, one may assume that having simply more customer and inspection data would help to detect NTL more accurately. However, in many cases, the data is biased as depicted in Fig.~\ref{fig:example}. 
Concretely, the customers inspected are a sample of the overall population of customers. In this example, there is a spatial bias. Hence, the inspections do not represent the overall population of customers. As a consequence, when learning from the inspection results, a bias is learned, making predictions less reliable. In technical terms, this bias is called covariate shift or sampling bias \cite{zadrozny2004learning}. Aside from spatial covariate shift, there may be other types of covariate shift in the data, such as the meter type, connection type, etc.

The main contributions of this paper are:
\begin{itemize}
\item We present a novel framework for quantifying and visualizing covariate shift at different hierarchical levels.
\item We investigate the importance of covariate shift to NTL detection in Big Data approaches and show that it leads to unreliable NTL predictors.
\item We report the covariate shift of different features in a real world data set consisting of 3.6M Brazilian customers and 820K inspection results.
\item We visualize how different local areas are affected by covariate shift at different hierarchical levels.
\end{itemize}
To the best of our knowledge, we are not aware of any previously published research that addresses this topic for NTL detection. We are convinced that an accurate study of this topic is necessary in order to advance NTL detection by reducing the covariate shift in data sets in the future.

%In general, the problem of covariate shift is ignored in most machine learning research. 
%The rest of this paper is organized as follows. Section \ref{chapter:review} provides a literature review of NTL detection and its challenges. Section \ref{chapter:covariateshift} defines covariate shift and describes the proposed methodology for measuring covariate shift. Section \ref{chapter:eval} presents experimental results, quantifications visualizations of the different types of covariate shift found in the data set. Section \ref{chapter:end} summarizes this work and provides an outreach on future work.

\section{Related Work}
\label{chapter:review}
From an electrical engineering perspective, energy balance methods \cite{oliveira2001new} can be applied to the detection of NTL. However, this requires topological information of the distribution network and does not reflect a change of network. In practice, the network topology undergoes rapid changes in emerging countries, i.e. the countries in which NTL is a particular issue.

The predominant methodology used in the literature is employing artificial intelligence (AI) to this anomaly detection problem. Historically, AI has been based on rule-based expert systems that incorporate expert knowledge. However, in many cases, describing a problem domain is challenging due to its complexity and temporal dynamics. Therefore, over the years, the fields of machine learning and data mining have become more popular. These methods learn models from data without being explicitly programmed.

%Profiles of 80K low-voltage and 6K high-voltage customers in Malaysia having meter readings every 30 minutes over a period of 30 days are used in \cite{nagi2010ntl} for electricity theft and abnormality detection. A test recall of 0.55 is reported.
A data set of \texttildelow 22K customers is used in \cite{costa2013fraud} for training a neural network. It uses the average consumption of the previous 12 months and other customer features such as location, type of customer, voltage and whether there are meter reading notes during that period. On the test set, an accuracy of 0.8717, a precision of 0.6503 and a recall of 0.2947 are reported.
Consumption profiles of 5K Brazilian industrial customer profiles are analyzed in \cite{ramos2012identification}. Each customer profile contains 10 features including the demand billed, maximum demand, installed power, etc. In this setting, a support vector machine slightly outperforms k-nearest neighbors and a neural network, for which test accuracies of 0.9628, 0.9620 and 0.9448, respectively, are reported. 

%\begin{itemize}
%\item Class imbalance and evaluation metric
%\item Feature description
%\item Incorrect inspection results
%\item Biased inspection results
%\item Scalability
%\item Comparison of different methods
%\end{itemize}

%Imbalanced classes in a data set refer to the property that it contains an unequal amount of labels per class. This fact is only addressed in very few NTL detection research. The imbalance also requires appropriate evaluation metrics that take this property into account. Feature description is a long-standing challenge in machine learning because learning algorithms often do not work on the raw data and need to be trained on features computed from the raw data. The set of inspected customers is a sample of all customers. This sample may not represent the overall population of customers as previous inspections have focused on certain areas. Furthermore, some inspection results reported are incorrect as technicians may have been threatened or bribed by fraudsters.

We have extensively reviewed the state of the art in our previous work and identified the open challenges in NTL detection \cite{glauner2016challenge}.
We have previously addressed the class imbalance and evaluation metric selection, when we showed that a large-scale machine learning approach outperformed rule-based Boolean and fuzzy logic expert systems \cite{glauner2016large}. Furthermore, we have shown that the neighborhood of customers yields significant information in order to decide whether a customer causes a NTL or not \cite{glauner2016neighborhood, meira2017distilling}.

\section{Covariate Shift}
\label{chapter:covariateshift}
In this paper, we address the challenge of covariate shift. It refers to the problem of training data (i.e. the set of inspection results) and production data (i.e. the set of customers to generate inspections for) having different distributions. This fact leads to unreliable NTL predictors when learning from this training data. In this section, we describe how to quantify it. 
Historically, covariate shift has been a long-standing issue in statistics. For example, The Literary Digest sent out 10M questionnaires in order to predict the outcome of the 1936 US Presidential election. They received 2.4M returns. Nonetheless, the predicted result proved to be wrong. The reason for this was that they used car registrations and phone directories to compile a list of recipients. In that time, the households that had a phone or a car represented a biased sample of the overall population. In contrast, George Gallup only interviewed 3K handpicked people, which were an unbiased sample of the population. As a consequence, Gallup could predict the outcome of the election very well \cite{harford2014big}.

\subsection{Definition}
Customers previously inspected are a sample of the overall population of customers. However, that sample may be biased, i.e. it does not represent the population of all customers, as visualized in Fig.~\ref{fig:example}. A reason for that is that previous inspections were largely focused on certain criteria and were not sufficiently spread among the population. 
The problem of training data and production data having different distributions has initially been addressed in the field of computational learning theory \cite{cortes2014domain}, which also calls it covariate shift, sampling bias or sample selection bias. It can be defined in mathematical terms as follows \cite{zadrozny2004learning}:
\begin{itemize}
\item Assume that all examples are drawn from a distribution $D$ with domain $X \times Y \times S$,
\item where $X$ is the feature space,
\item $Y$ is the label space,
\item and $S$ is $\{0,1\}$.
\end{itemize}

Examples $(x, y, s)$ are drawn independently from $D$. $s=1$ denotes a selected example, whereas $s=0$ denotes the opposite. The training is performed on a sample that comprises all examples that have $s = 1$. $P(s\lvert x, y) = P(s\lvert x)$ implies that $s$ is independent of $y$ given $x$. In this case, the selected sample is biased but the bias only depends on the feature vector $x$. This problem is called covariate shift \cite{zadrozny2004learning}.

\subsection{Big Data}
In the past fifteen years, the Big Data paradigm followed can be summarized: ``It's not who has the best algorithm that wins. It's who has the most data." \cite{banko2001scaling} 
Concretely, this approach reflects the desire of deriving better conclusions from simply analyzing more data, without the necessity of looking at theory and models. 
However, taking covariate shift into account, just having more data is not sufficient. In this case, having less data that is more representative seems to be the answer.

\subsection{Affected Classifiers}
It has been shown that some machine learning algorithms are not affected by covariate shift, whereas others are \cite{zadrozny2004learning}.

\subsubsection{Local learner}
The prediction of the learner depends asymptotically only on $P(y\lvert x)$. Hence, it is not affected by covariate shift. Examples are logistic regression and hard-margin support vector machine (SVM).
\subsubsection{Global learner}
The prediction of the learner depends asymptotically on both, $P(y\lvert x)$ and $P(x)$. Hence, it is affected by covariate shift.
Example: Decision tree learners such as ID3 or C4.5 \cite{quinlan1993c4} recursively split the input space by choosing the remaining most discriminative feature of a data set. To predict, the learned tree is traversed top-down. Other examples are naive Bayes and soft-margin SVM.

\subsection{Quantification}
\label{chapter:covariateshift:quant}
The Kullback-Leibler divergence is a measure of the difference of two probability distributions. However, it is challenging (1) to adapt this measure to multi-dimensional data that is a combination of discrete and continuous features, which is common in machine learning, and (2) to define criteria when a distance is an indicator for a covariate shift.

Therefore, a preferred methodology for quantifying covariate shift is: First, we add a feature \texttt{s} and assign the values \texttt{1} or \texttt{0} to the training data $(s = 1)$ or production data $(s = 0)$, respectively. 
These data sets are furthermore merged into one data set. This latter is split into a training set \texttt{X1} (with no relation to the original training set) and a test set \texttt{X2}.
The objective is to develop a supervised learning method capable of predicting the feature \texttt{s} using \texttt{X1}.
The performance of the classifier on \texttt{X2} is then quantified using the Matthews correlation coefficient (MCC)
\begin{align}
 \frac{TP\times TN - FP\times FN}{\sqrt{(TP + FP)(TP+FN)(TN+FP)(TN+FN)}},
\end{align}
which measures the accuracy of binary classifiers taking into account the imbalance of both classes, ranging from $-1$ to $+1$ \cite{matthews1975comparison}.
The greater the MCC, the greater the covariate shift. A concrete threshold for covariate shift depends on the problem, however 0.2 has been proposed \cite{bigml2014covarite}. Though a low MCC does not automatically imply the lack of a covariate shift, a significant MCC value is an indicator of covariate shift. 

%It uses the number of true positives (TP), false positives (FP), true negatives (TN) and false negatives (FN) for this calculation.

We extend this methodology in Algorithm~1 by introducing the following novelties:
\begin{enumerate}
\item Tree classifier: Decision tree learning is affected by covariate shift. Decision trees scale to very large data sets while they allow to learn non-linearities. Soft-margin SVMs are also global learners, however, for large data sets only a linear kernel is learnable in a feasible amount of time.
\item Model selection: We want to find a model which is able to distinguish between both distributions. 
Thus maximizing the MCC on the test set is equivalent to finding the best two-class classification between production data and original training data.
For this, we optimize the five most important tree model parameters by randomly drawing from probability distributions: Max. number of leaves, max. number of levels, measure of the purity of a split, min. number of samples required to be at a leaf and min. number of samples required to split a node.
\item Cross-validation: We also split the data set into $k$ folds in order to reduce the overfitting. This leads to a more reliable model for covariate shift quantification. The MCC per model, denoted by $\overline{MCC}$, is the average of the MCCs of the $k$ test sets. The standard deviation of the $k$ test MCCs serves as the reliability of $\overline{MCC}$. The lower the standard deviation, the more reliable $\overline{MCC}$.
\end{enumerate}
Note: The inspection results are not taken into account as covariate shift only concerns the distributions of the inputs.

\begin{algorithm}
\begin{algorithmic}[1]
\State $result \gets 0$
\State $reliability \gets 0$
\State $selected \gets train\_data.add\_feature(s, 1)$
\State $not\_selected \gets prod\_data.add\_feature(s, 0)$
\State $data \gets selected \cup not\_selected$
\State $folds \gets cv\_folds(data, k)$
\For{$model$ \textbf{in} $get\_model\_candidates()$}
\State $mccs \gets list()$
\For{$fold$ \textbf{in} $folds$}
\State $X_{train}, X_{test}, y_{train}, y_{test} \gets fold$
\State $classifier \gets DecisionTree(model)$
\State $classifier.train(X_{train}, y_{train})$
\State $y_{pred} \gets classifier.predict(X_{test})$
\State $mccs.append(MCC(y_{test}, y_{pred}))$
\EndFor
\State $mcc\_mean \gets mean(mccs)$
\If{$mcc\_mean > result$}
\State $result \gets mcc\_mean$
\State $reliability \gets std(mccs)$
\EndIf
\EndFor
\State \textbf{return} $result, reliability$
\end{algorithmic}
\label{alg:mcc}
\caption{Quantifying covariate shift.}
\end{algorithm}

\section{Evaluation}
\label{chapter:eval}
\subsection{Data}
The data used in this paper comes from an electricity provider in Brazil. First, it consists of 3.6M customers. A complete list of the customer master data used in the following experiments is depicted in Table~\ref{table:features}. The categorical features class, voltage, number of wires, contract status and meter type are converted to one-hot coding.
Second, the data contains 820K inspection results, such as inspection date, presence of fraud or irregularity, type of NTL and inspection notes. 620K customers have been inspected at least once and the remaining \texttildelow 3M customers have never been inspected.
Third, there are 195M meter readings from 2011 to 2016 such as consumption in kWh, date of meter reading and number of days between meter readings.

\begin{table}[!t]
\renewcommand{\arraystretch}{1.3}
% if using array.sty, it might be a good idea to tweak the value of
% extrarowheight as needed to properly center the text within the cells
\caption{Assessed Features.}
\label{table:features}
\centering
\begin{tabular}{|l|c|}
\hline
Feature & Possible values\\
\hline
\hline
Class & Power generation infrastructure, residential, \\
& commercial, industrial, public, \\
& public illumination, rural, public service, \\
& reseller \\
\hline
Contract status & Active, suspended \\
\hline
Location & Longitude and latitude \\
\hline
Meter type & 22 different meter types \\
\hline
Number of wires & 1, 2, 3 \\
\hline
Voltage & $\le$2.3kV, $>$2.3kV \\
\hline
\end{tabular}
\end{table}

\subsection{Implementation Notes}
All computations were run on a server with 24 cores and 128 GB of RAM. 
The entire code was implemented in Python using \texttt{scikit-learn} \cite{scikit-learn} for machine learning. \texttt{scikit-learn} allows to distribute the training of the cross-validated classifiers among all cores. The maps were plotted using \texttt{cartopy} \cite{Cartopy}. In total, all results and plots reported in this paper were computed in 12 hours using this infrastructure. Our implementation is available as open source: \texttt{http://github.com/pglauner/SpatialBiasNTL}.

\subsection{Model Parameters}
In the following experiments, we use $k=10$-fold cross-validation. In each experiment, we train 1K trees, which are 100 different tree models trained on each of the 10 folds. We optimize the five tree model parameters by randomly drawing from predefined uniform probability distributions depicted in Table~\ref{table:params}. We have chosen these ranges based on best practice recommendations and our own experience.
Furthermore, the two classes ($s=1$ and $s=0$) are imbalanced, i.e. there are more examples of the non-inspected customers than the inspected ones. In order to take this into account during training, we associate weights with the classes such that the examples of the minority class have stronger impact.

\begin{table}[!t]
\renewcommand{\arraystretch}{1.3}
% if using array.sty, it might be a good idea to tweak the value of
% extrarowheight as needed to properly center the text within the cells
\caption{Tree Model Parameters.}
\label{table:params}
\centering
\begin{tabular}{|l|c|}
\hline
Parameter & Range \\
\hline
\hline
Max. number of leaves & $[2, 20)$ \\
\hline
Max. number of levels & $[1, 20)$ \\
\hline
Measure of the purity of a split & $\{$entropy, gini$\}$ \\
\hline
Min. number of samples required to be at a leaf & $[1, 20)$ \\
\hline
Min. number of samples required to split a node & $[2, 20)$ \\
\hline
\end{tabular}
\end{table}

\subsection{Global Covariate Shifts}
In the following experiments, we compute different global types of covariate shift by using all customers in each experiment. We therefore do not split the customers into different geographical areas.

We have previously presented the customer master data features available in Table~\ref{table:features}. We compute the global covariate shift of each of these features. We report our results in Table~\ref{table:mcc_global1}. $\overline{MCC}_{max}$ denotes the maximum average of the MCCs of the $k=10$ test sets among all 100 tree models trained on a feature. $\sigma$ denotes the standard deviation of those $k=10$ MCC test scores, which is a reliability measure of $\overline{MCC}_{max}$.

\begin{table}[!t]
\renewcommand{\arraystretch}{1.3}
% if using array.sty, it might be a good idea to tweak the value of
% extrarowheight as needed to properly center the text within the cells
\caption{Global Covariate Shift of Single Features.}
\label{table:mcc_global1}
\centering
\begin{tabular}{|l|c|c|}
\hline
Feature & $\overline{MCC}_{max}$ & $\sigma$\\
\hline
\hline
Location & \textbf{0.22367} & 0.03453 \\
\hline
Class & 0.16255 & 0.01371 \\
\hline
Number of wires & 0.14111 & 0.00794 \\
\hline
Meter type & 0.13158 & 0.00382 \\
\hline
Voltage & 0.07092 & 0.02375 \\
\hline
Contract status & 0.03744 & \textbf{0.09183} \\
\hline
\end{tabular}
\end{table}

\begin{figure}[!t]
\centering
\includegraphics[width=0.4\textwidth]{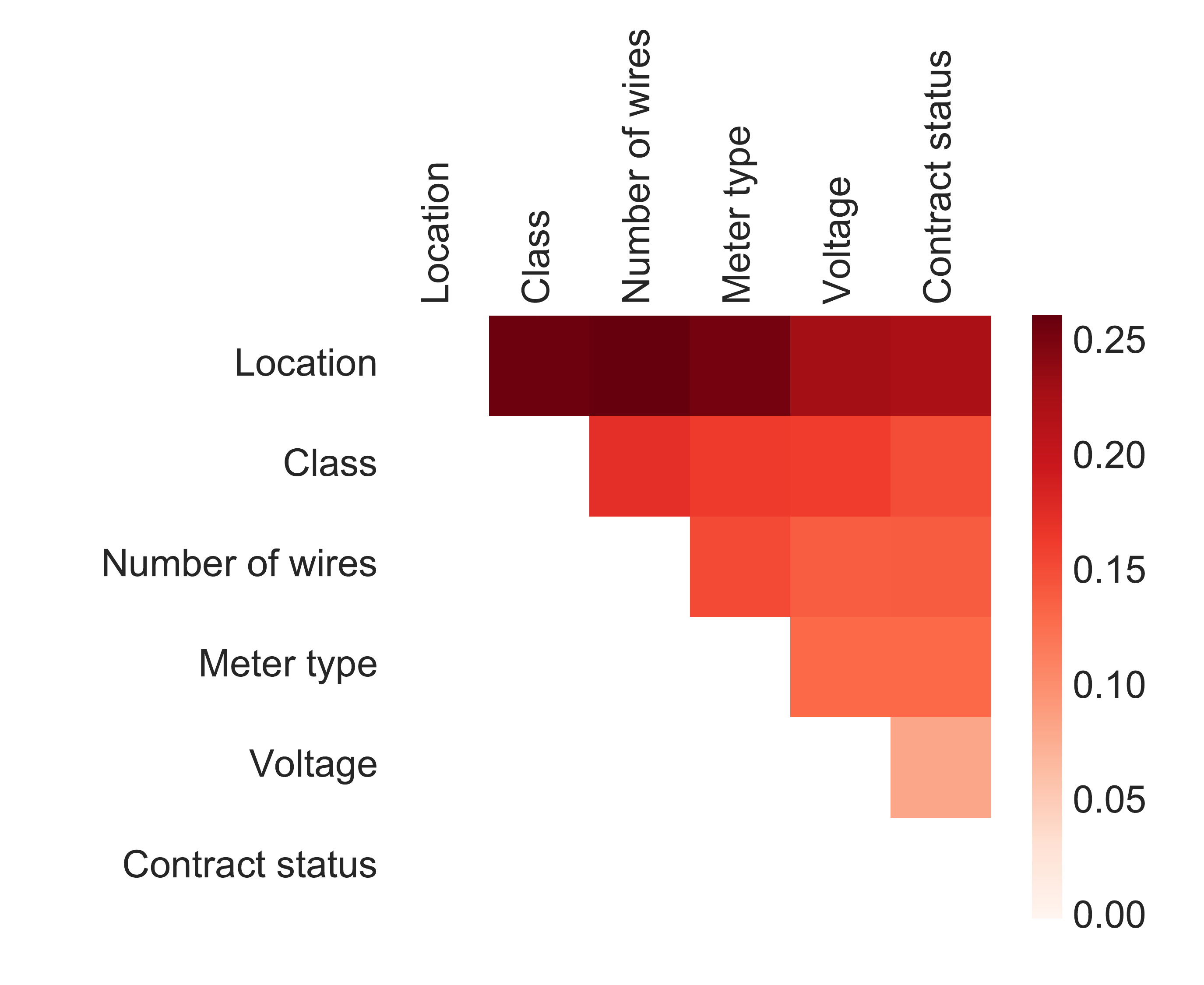}
\caption[XXX]{Global Covariate Shift of Compound Features.}
\label{fig:mcc_global2}
\end{figure}

Overall, the strongest covariate shift is in the location with a MCC value of 0.22367. This means that previous inspections are mostly biased towards the location of customers. The location is also the only feature that is beyond the threshold of 0.2 mentioned in Section~\ref{chapter:covariateshift:quant}. The features class, number of wires and meter type are below the threshold but are greater than 0.1. There is almost no covariate shift of previous inspections towards the voltage and contract status features.
The standard deviation of the MCCs is the greatest for the contract status. The reason for this is a strong overfit in one of the folds. All other MCCs have a much lower standard deviation, making them more reliable.

Next, we create compound features that are composed of multiple features. Due to the great number of possible combinations, we assess all 2-combinations as well as the 6-combination of all features. We visualize the MCCs for all 2-combinations in Fig.~\ref{fig:mcc_global2} and reported the MCCs in Table~\ref{table:mcc_global2}. For the 6-combination comprising all features, we computed $\overline{MCC}_{max}=0.27325$, which is the maximum covariate shift of all compound features. Therefore, the spatial covariate shift contributes to this covariate shift the most, however, the other covariate shifts contribute a fraction as well.

\begin{table}[!t]
\renewcommand{\arraystretch}{1.3}
% if using array.sty, it might be a good idea to tweak the value of
% extrarowheight as needed to properly center the text within the cells
\caption{Global Covariate Shift of Compound Features.}
\label{table:mcc_global2}
\centering
\begin{tabular}{|l|c|c|}
\hline
Feature & $\overline{MCC}_{max}$ & $\sigma$\\
\hline
\hline
All & \textbf{0.27325} & 0.03014 \\
\hline
\hline
Location + number of wires & \textbf{0.26206} & 0.03676 \\
\hline
Location + class & 0.25796 & 0.03540 \\
\hline
Location + meter type & 0.25479 & 0.03884 \\
\hline
Location + voltage & 0.22944 & 0.03544 \\
\hline
Location + contract status & 0.22335 & 0.03454 \\
\hline
Class + number of wires & 0.17501 & 0.00468 \\
\hline
Class + meter type & 0.16472 & 0.00309 \\
\hline
Class + voltage & 0.16322 & 0.01400 \\
\hline
Number of wires + meter type & 0.15283 & 0.00274 \\
\hline
Class + contract status & 0.15158 & 0.00992 \\
\hline
Number of wires + voltage & 0.14156 & 0.00800 \\
\hline
Number of wires + contract status & 0.14111 & 0.00794 \\
\hline
Meter type + voltage & 0.13165 & 0.00381 \\
\hline
Meter type + contract status & 0.13155 & 0.00382 \\
\hline
Voltage + contract status & 0.08213 & \textbf{0.08301} \\
\hline
\end{tabular}
\end{table}

\subsection{Local Covariate Shifts}
We now entirely focus on spatial covariate shift since it is the strongest one among the different types of covariate shift. In the following experiments we compute local covariate shifts by splitting the customers in different locations. The data set provides the following divisions in the following hierarchical order:
\begin{enumerate}
\item 9 regions
\item 261 municipalities
\item 1,380 localities
\item 19,026 neighborhoods
\end{enumerate}

\begin{figure}
  \subfloat[Regional level.]{\label{fig:local:a} \includegraphics[width=0.5\textwidth]{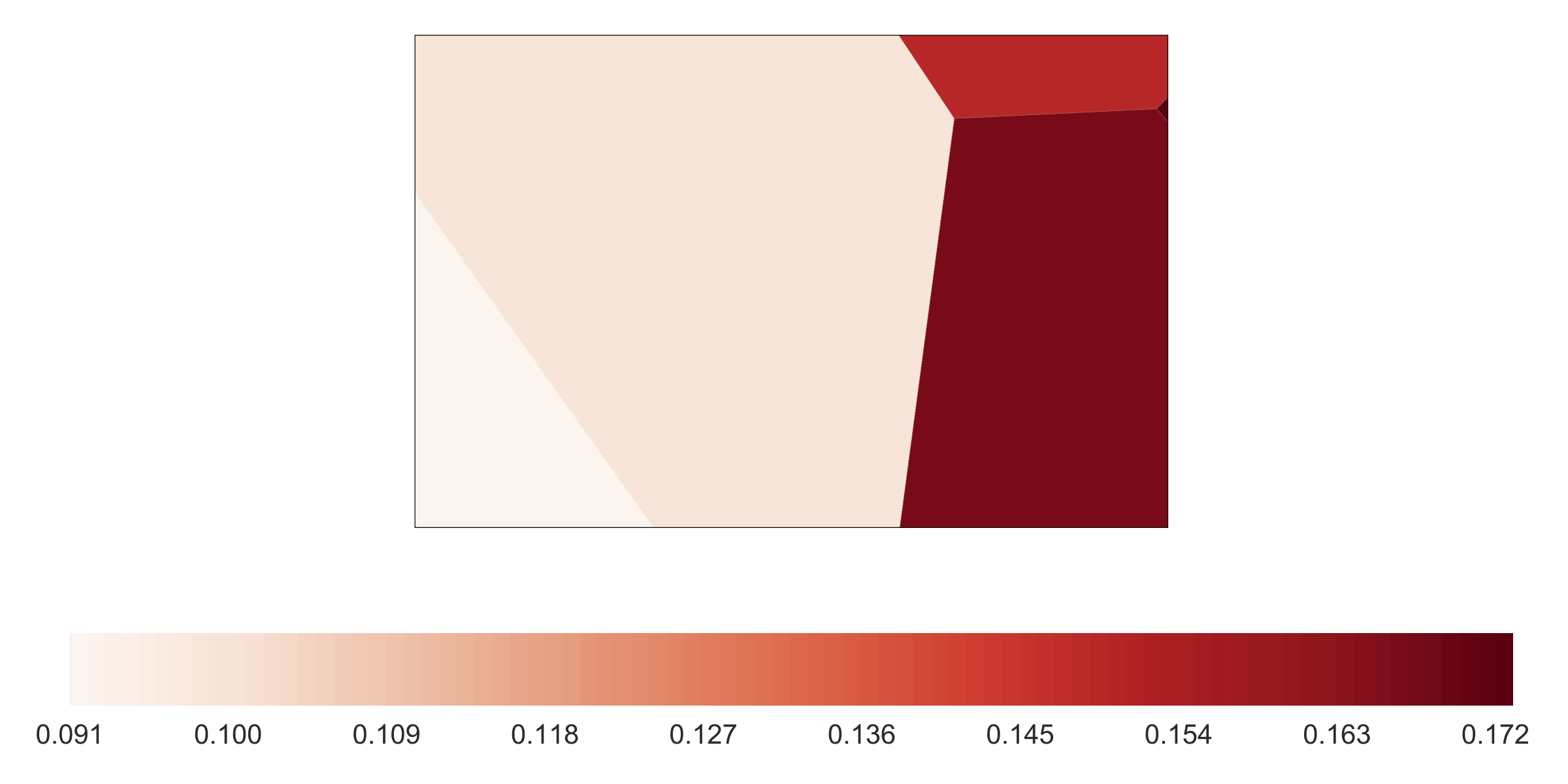}}\\
  \subfloat[Municipal level.]{\label{fig:local:b} \includegraphics[width=0.5\textwidth]{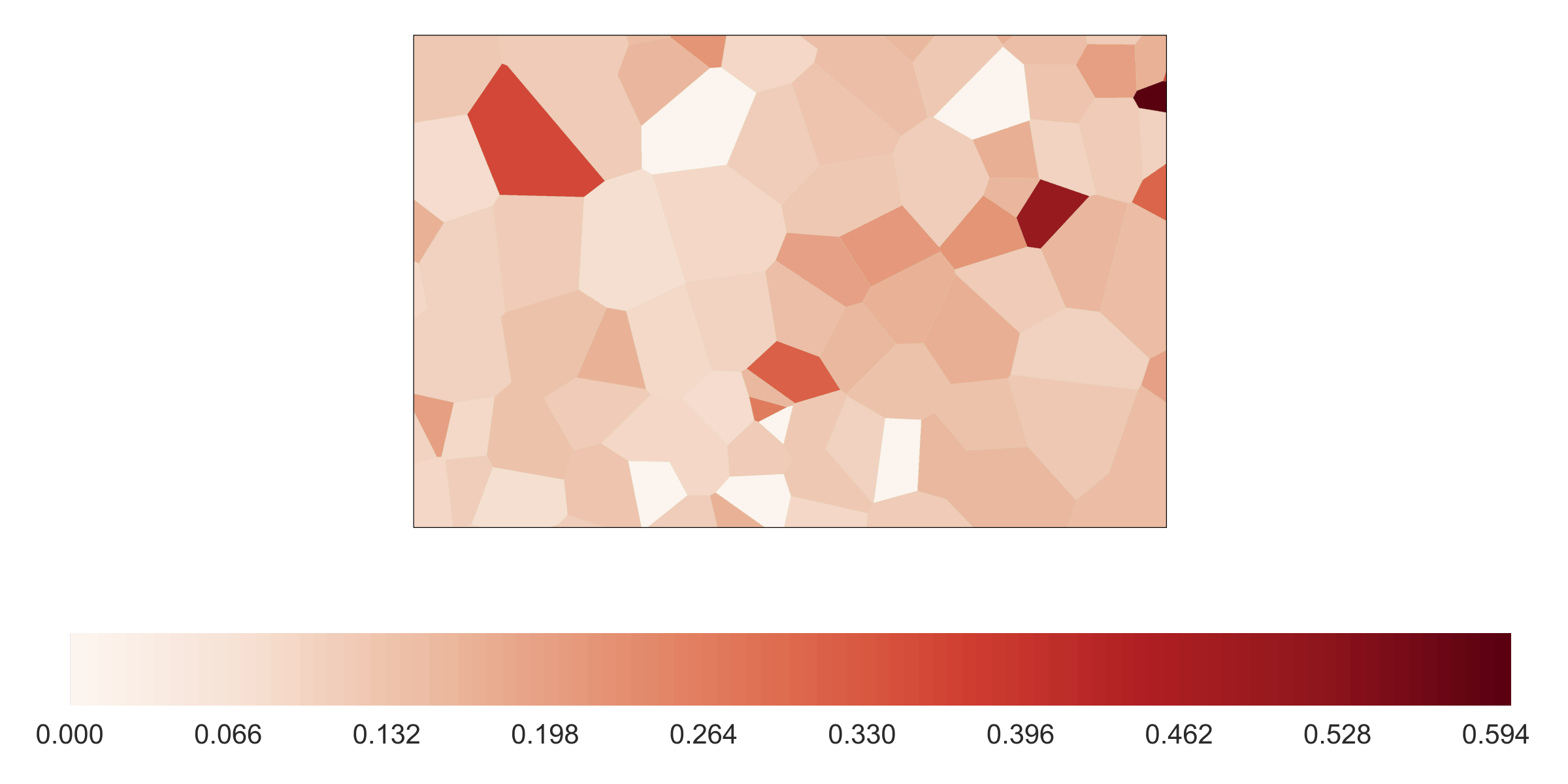}}\\
  \subfloat[Local level.]{\label{fig:local:c} \includegraphics[width=0.5\textwidth]{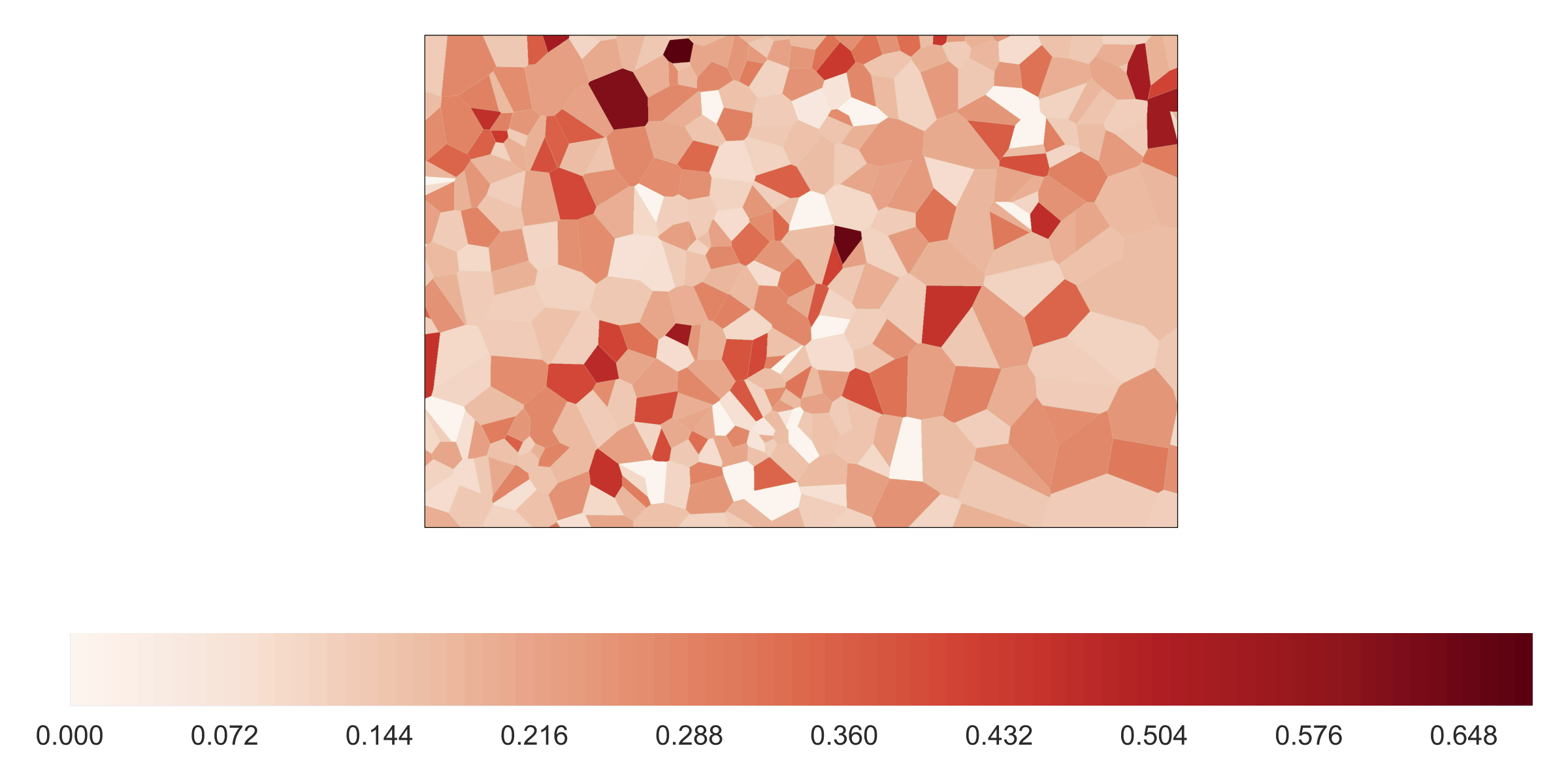}}\\
  \subfloat[Neighborhood level.]{\label{fig:local:d} \includegraphics[width=0.5\textwidth]{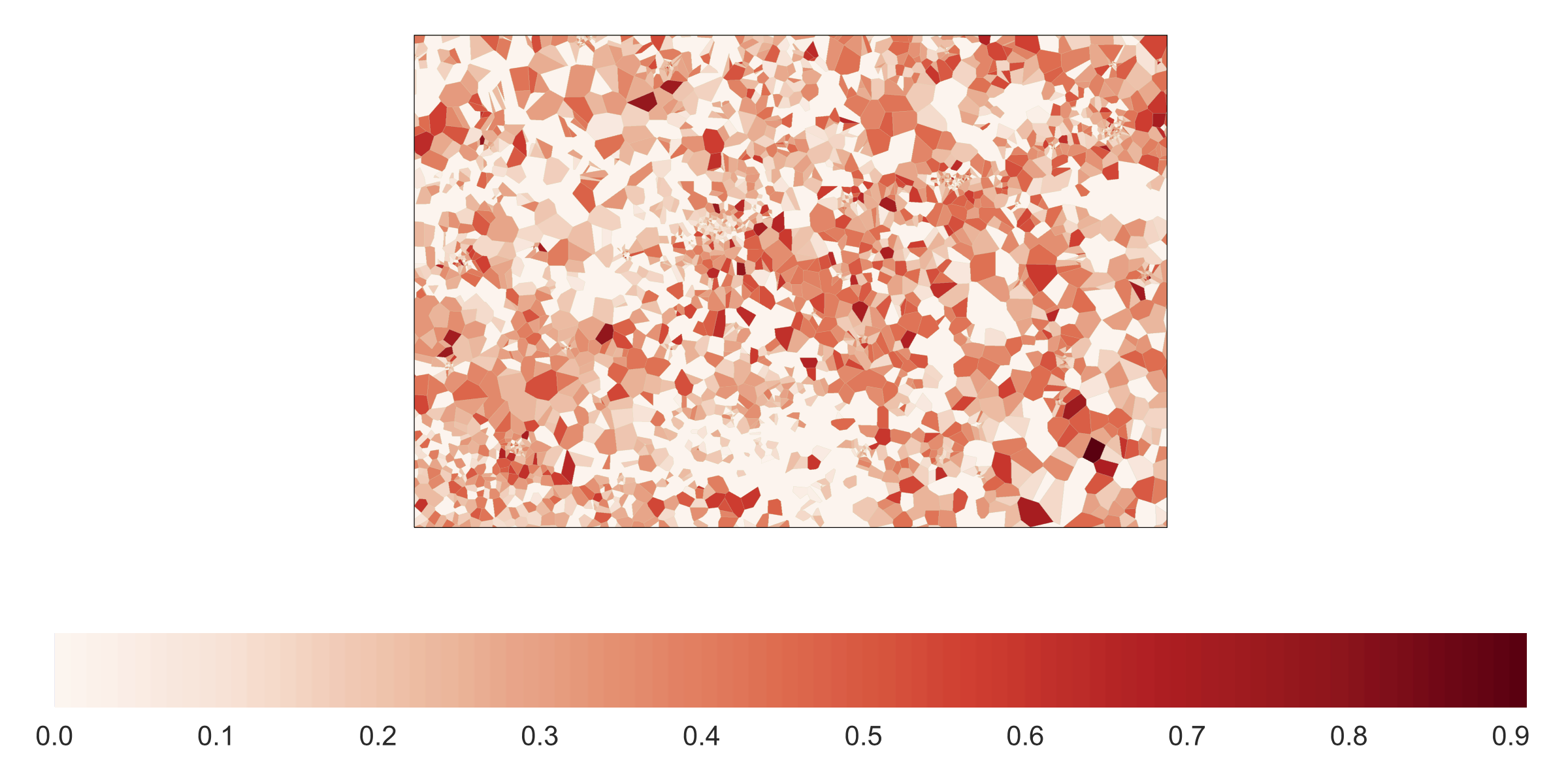}}\\
\caption{Spatial covariate shift at different levels of granularity. For each  division, we compute the median location of the respective customers and assign $\overline{MCC}_{max}$ to it. We then use nearest interpolation to generate the local covariate shift maps. Note: The maps only visualize a fraction of all political divisions. This is because we have cropped the computed maps in this paper due to privacy purposes.}
\label{fig:local}
\end{figure}

We plot the local spatial covariate shifts of these four levels of granularity in Fig.~\ref{fig:local}. All customers are located in one Brazilian state.
We observe that spatial covariate shift is smoothened for regional level in Fig.~\ref{fig:local:a}. It becomes increasingly more granular at municipal, local and neighborhood levels in Figs.~\ref{fig:local:a} through \ref{fig:local:d}, respectively. We also notice that the spatial covariate shifts at lower levels tend to increase, which is depicted by increasing upper limits of the color bars.

\subsection{Discussion}
We have shown that covariate shift exists in our real world data set. The features of the customer data that are most affected by covariate shift are the location, followed by class, number of wires and meter type. Classifiers that use other features such as the voltage or contract status instead tend to be more reliable.
We have also shown that the spatial covariate shift exists on different levels of granularity and that municipalities and localities with very strong covariate shifts exist. Subsequently, these local covariate shifts have significant impact on the covariate shifts on higher levels or even globally on the entire data set. Therefore, using all inspection results of the data set for training a NTL predictor from a Big Data perspective leads to biased models that may not reliably detect NTL. As a consequence, reducing the spatial covariate shift in the data set must be a priority in order to learn reliable NTL predictors.

However, the finer the hierarchical granularity, the more divisions cannot be used for the computations for the following reasons. First, using $k=10$-fold cross-validation, training is only possible if a division has at least $k$ customers. Second, the $k-1$ folds used for training must have examples of both classes.
Third, the MCC can only be computed for denominator $\ne 0$, which is the case for $(TP > 0 \wedge TN > 0) \vee (FP > 0 \wedge FN > 0)$. If the test MCC cannot be computed for a fold of a model, only the MCCs of the remaining folds are used in cross-validation.
If no MCCs can be computed for a division, we skip it in the plotting. For instance, this effect has become most apparent at neighborhood level in the west of that state due to the low population density.

\section{Conclusion and Future Work}
\label{chapter:end}
Covariate shift is the problem of training data and production data having different distributions. In this work, we have proposed a novel framework for quantifying and visualizing covariate shift in spatial data sets. In the context of non-technical loss (NTL) detection, we showed that there is a covariate shift between the inspected customers and the overall population of customers. We used a real world data set from Brazil that consists of 3.6M customers and 820K inspection results. We showed that some features have a stronger covariate shift than others. In particular, the spatial covariate shift is the strongest and appears in different hierarchical levels. Subsequently, machine learning models trained on this data will lead to unreliable NTL predictions.
Our contribution will allow domain experts to model more reliable rules for NTL predictors.
%The algorithms for measuring and visualizing covariate shift are about to be deployed in a CHOICE Technologies product that uses an expert system for NTL detection.

Next, we will use and amend spatial point processes \cite{baddeley2007spatial} for reducing the spatial covariate shift in our data set in order to train more reliable NTL predictors.
%Spatial point processes build on top of Poisson processes and allow to examine a data set of spatial locations and to conclude whether the locations are randomly distributed in a space or if they are skewed.

\section*{Acknowledgment}
We would like to thank Lautaro Dolberg, Diogo Duarte and Yves Rangoni from CHOICE Technologies Holding S\`arl for contributing many good ideas to our discussions. This work has been partially funded by the Luxembourg National Research Fund.



\bibliographystyle{IEEEtran}
\bibliography{references}

\end{document}